# Tracking the Spatiotemporal Evolution of Landslide Scars Using a Vision Foundation Model: A Novel and Universal Framework

Meijun Zhou [a], Gang Mei [a,*], Zhengjing Ma [a], Nengxiong Xu [a], Jianbing Peng [a,b]

Corresponding author: gang.mei@cugb.edu.cn (Gang Mei)

[a] School of Engineering and Technology, China University of Geosciences (Beijing), 100083, Beijing, China

[b] School of Geological Engineering and Geomatics, Chang'an University, 710064, Xi'an, China

**Abstract**

Tracking the spatiotemporal evolution of large-scale landslide scars is critical for understanding the evolution mechanisms and failure precursors, enabling effective early-warning. However, most existing studies have focused on single-phase or pre- and post-failure dual-phase landslide identification. Although these approaches delineate post-failure landslide boundaries, it is challenging to track the spatiotemporal evolution of landslide scars. To address this problem, this study proposes a novel and universal framework for tracking the spatiotemporal evolution of large-scale landslide scars using a vision foundation model. The key idea behind the proposed framework is to reconstruct discrete optical remote sensing images into a continuous video sequence. This transformation enables a vision foundation model, which is developed for video segmentation, to be used for tracking the evolution of landslide scars. The proposed framework operates within a knowledge-guided, auto-propagation, and interactive refinement paradigm to ensure the continuous and accurate identification of landslide scars. The proposed framework was validated through application to two representative cases: the post-failure Baige landslide and the active Sela landslide (2017–2025). Results indicate that the proposed framework enables continuous tracking of landslide scars, capturing both failure precursors critical for early warning and post-failure evolution essential for assessing secondary hazards and long-term stability.

**Keywords:** Large-Scale Landslides, Landslide Scars, Spatiotemporal Evolution, Optical Remote Sensing, Vision Foundation Model

## 1. Introduction

Large-scale landslides, one of the most destructive geological hazards worldwide, pose severe threats to human safety and regional infrastructure (Fan et al., 2017; Zhang et al., 2018; Fan et al., 2019). Therefore, identifying and monitoring large-scale landslides is essential for early disaster detection, risk assessment, and the development of effective prevention and mitigation strategies.

Landslides are not instantaneous events but dynamic processes that evolve through gradual accumulation, progressive deformation, and eventual failure (Pánek et al., 2016; Li et al., 2023; Sun et al., 2024; Li et al., 2025). A critical challenge in landslide research lies in accurately capturing their spatiotemporal evolution, with particular emphasis on the dynamic development of landslide scars as direct surface evidence of activity. **Landslide scars**, defined as the zones where the original ground surface has been displaced or removed by downslope movement of soil, rock, or debris, provide essential insights into ongoing processes (Rana et al., 2021; Bhuyan et al., 2024; Bhuyan et al., 2025). This evolution involves phenomena such as the expansion of affected areas, boundary merging, and variations in pre-failure movement (Hu et al., 2020; Urgilez Vinueza et al., 2022). Understanding these processes is fundamental for elucidating landslide evolution mechanisms, identifying evolutionary stages, recognizing potential precursors to failure, and establishing effective early warning systems (Lacroix et al., 2023; Mei et al., 2025).

Landslide scars are primarily identified through ground-based monitoring and remote sensing approaches, such as unmanned aerial vehicle (UAV) photogrammetry and satellite imagery (Casagli et al., 2023). Although ground-based and UAV methods provide high precision, their applicability to long-term, large-scale, and continuous monitoring is limited (Ma et al., 2021; Casagli et al., 2023). By contrast, remote sensing has become an indispensable tool for investigating the spatiotemporal evolution of large-scale landslide scars due to its wide spatial coverage and frequent revisit capability (Ma and Mei, 2021; Casagli et al., 2023). In particular, synthetic aperture radar interferometry (InSAR) has proven highly effective for monitoring slow-moving landslides. Nevertheless, InSAR performance is often constrained by decoherence resulting from dense vegetation, steep terrain, and high deformation gradients, which hinders continuous tracking of landslide evolution, especially once the slope enters an accelerated deformation phase (Yao et al., 2017; Zhang et al., 2018; Bekaert et al., 2020;

Xiong et al., 2020; Casagli et al., 2023).

Extensive research has demonstrated that catastrophic landslides typically experience a progressive deformation stage prior to failure, characterized by crack propagation, vegetation degradation, and gradual expansion of bare ground (Fan et al., 2019; Yang et al., 2019; Guo et al., 2021; Li et al., 2023). These deformation features, manifested in the evolution of landslide scars, can be effectively detected and quantified through multi-temporal optical imagery over timescales ranging from several years to decades (Yang et al., 2019; Yang et al., 2020; Wang et al., 2023). Traditional approaches predominantly employ manual interpretation and multi-temporal index comparisons, such as the normalized difference vegetation index (NDVI), to delineate landslide boundaries. However, these approaches are inefficient and inadequate for capturing dynamic changes in large-scale landslide scars over extended time-series because they inherently reduce continuous evolution processes to analyses at discrete time points (Yang et al., 2019; Li et al., 2023).

In recent years, deep learning approaches, particularly convolutional neural networks (CNNs), have substantially enhanced the efficiency and accuracy of spatial landslide identification (Wang et al., 2019; Liu et al., 2021; Ullo et al., 2021; Meena et al., 2022; Lv et al., 2023; Wan et al., 2023). The majority of applications are devoted to single-temporal object detection and change detection between pre- and post-event imagery. More recently, the advent of vision foundation models (VFMs), such as the Segment Anything Model (SAM), has introduced a new paradigm, demonstrating considerable potential to segment landslide boundaries from single-temporal imagery via zero-shot learning capabilities (Kirillov et al., 2023; Hou et al., 2025; Yang et al., 2025). Building on this potential, researchers have further adapted and refined SAM for landslide detection through advanced techniques, such as multimodal fusion and prompt engineering (Yu et al., 2024; He et al., 2025; Hou et al., 2025; Wang et al., 2025; Yang et al., 2025). However, despite their power in spatial feature extraction, these advanced models remain bound to a static paradigm. Consequently, these models struggle to continuously track the evolution of landslide scars in a frame-by-frame manner, as is done in video analysis.

In summary, both traditional manual interpretation and most existing deep learning methods for landslide identification are predominantly rely on a static analysis paradigm. This paradigm involves analyzing discrete time points (e.g., single-phase imagery or dual-phase imagery representing pre- and post-failure conditions) to delineate the extent and morphology of landslides. Lacking the capacity to treat long-term optical imagery as a continuous dynamic process, these approaches struggle to capture key dynamic features, such as the progressive expansion of boundaries or periodic variations in activity. This limitation restricts the potential to deeply understand landslide evolution mechanisms and accurately identify failure precursors.

To address the aforementioned problem, this study proposes a novel and universal framework for tracking landslide scars using a vision foundation model. In contrast to previous approaches, the proposed framework continuously captures the spatiotemporal evolution of landslide scars before and after failure by reformulating discrete multi-temporal remote sensing image analysis as a video segmentation problem. The proposed framework operates under a knowledge-guided, auto-propagation, and interactive refinement paradigm. It incorporates domain knowledge (prompt points) in the initial frame and automatically tracks and segments landslide scars in subsequent frames. Moreover, when the identification results show substantial discrepancies from the ground truth, an interactive refinement mechanism enables corrections by adding additional prompt points. To evaluate the effectiveness of the proposed framework, two scenarios, i.e., post-failure landslides and active landslides, were used to track the spatiotemporal evolution of landslide scars at different stages.

## 2. Method

### 2.1 Overview of the Framework

This study aims to develop a universal framework for automatically and continuously tracking landslide scars from multi-temporal optical remote sensing imagery. The framework is based on the concept of reformulating discrete image recognition tasks as continuous video segmentation tasks. By integrating an advanced vision foundation model, the framework leverages the model's robust spatiotemporal feature learning and segmentation capabilities to capture the evolution of landslide scars. The workflow of the proposed framework consists of three main steps, as illustrated in Fig. 1.

(1) Acquiring and preprocessing optical remote sensing imagery: Multi-temporal, multi-spectral optical remote sensing images were acquired and subjected to a series of preprocessing steps to generate standardized index images, such as the NDVI, which effectively capture changes in surface vegetation and landforms. These standardized images provide consistent and reliable data for subsequent identification of landslide scars.

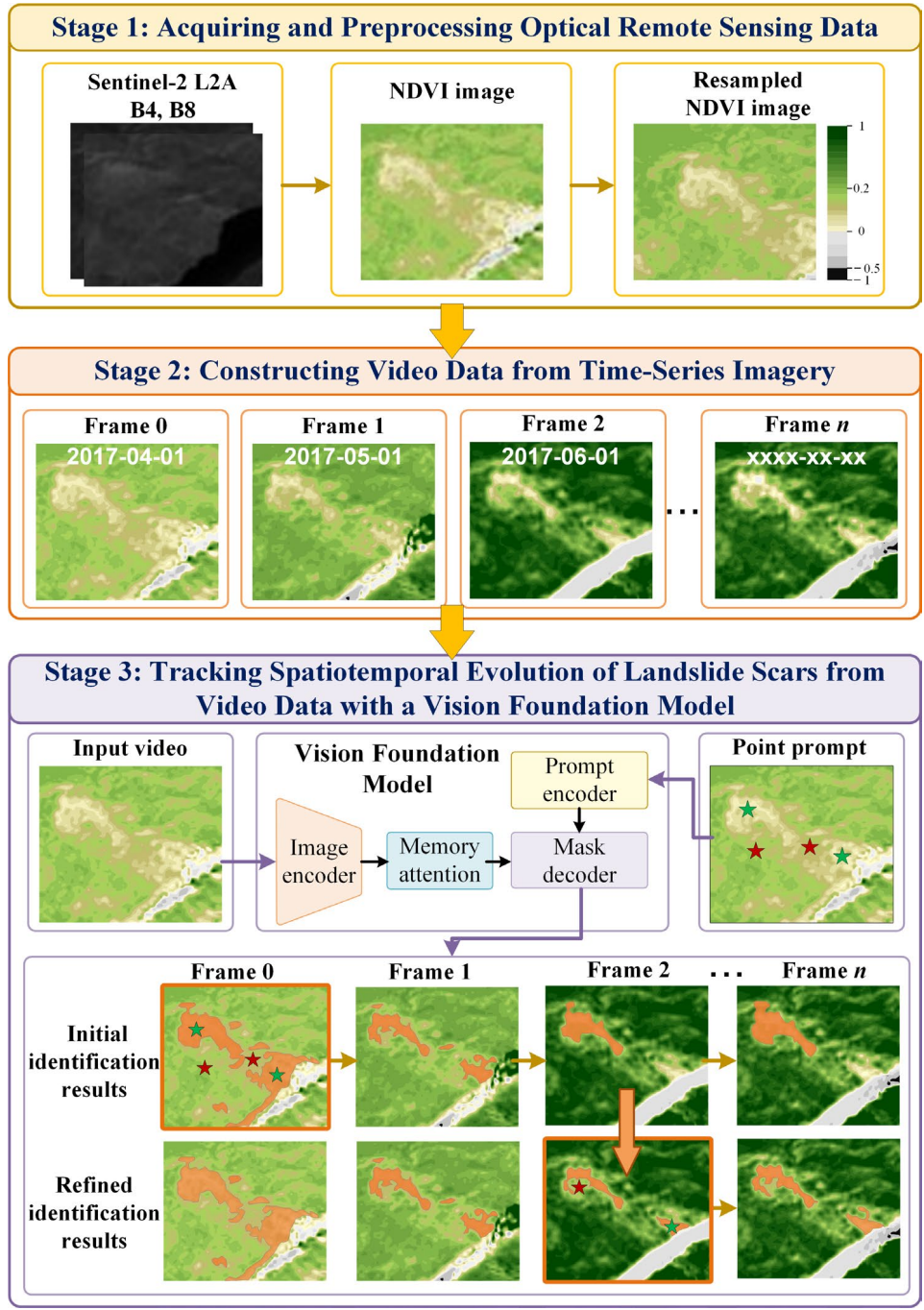

**Fig. 1.** Workflow of the proposed universal framework for tracking the spatiotemporal evolution of landslide scars.

(2) Constructing video data from time-series imagery: Preprocessed time-series index images are organized as temporal video sequences. This transformation converts originally discrete, static landslide images into a continuous dynamic format, enabling continuous identification of landslide scars throughout their evolution.

(3) Tracking the spatiotemporal evolution of landslide scars using a vision foundation model: The constructed video sequences are processed by an advanced vision foundation model under a knowledge-guided, auto-propagation, and interactive refinement framework. First, the landslide scar is initialized by incorporating limited domain knowledge (prompt points) into the initial frame. The model then automatically tracks and segments landslide scars in subsequent frames via its temporal memory mechanism. Finally, an interactive refinement mechanism enables corrections at any intermediate frame, ensuring continuity and accuracy in the final results.

**2.2 Workflow of the Proposed Framework**

2.2.1 Stage 1: Acquiring and Preprocessing Optical Remote Sensing Data

This study uses atmospheric-corrected Level-2A (L2A) products from the Sentinel-2 satellite as the primary data source. Progressive deformation of large-scale landslides is often accompanied by vegetation degradation and expansion of bare ground. Therefore, the NDVI, which is particularly sensitive to changes in vegetation cover, was chosen as the key indicator for identifying landslide scars. The NDVI highlights areas affected by vegetation damage while reducing the influence of mountain shadows (Fiorucci et al., 2019; Yang et al., 2019). NDVI values were calculated using Eq. (1).

$$NDVI = \frac{(\rho_{nir} - \rho_r)}{(\rho_{nir} + \rho_r)} \qquad (1)$$

where $\rho_{nir}$ and $\rho_r$ represent the reflectance values in the near-infrared and red bands, respectively. Using true-color and false-color imagery of the study area for the same period, along with information from previous studies (Yang et al., 2019; Guo et al., 2021), the optimal NDVI threshold was determined statistically for the study area. This threshold was subsequently applied to identify landslide scars, thereby enabling quantitative tracking of their spatiotemporal evolution.

To enhance the accuracy of the landslide scar boundary identification and mitigate edge artifacts in the original 10 m resolution imagery, bilinear interpolation was employed to resample the NDVI time-series images. Based on the spatial scales of the study areas, the imagery was resampled to either 1 m or 2 m resolution to optimize the trade-off between spatial detail and computational efficiency. A comparison of the imagery before and after resampling is shown in Fig. 2. Finally, all temporal images were precisely cropped and registered to ensure a consistent spatial reference for subsequent analysis.

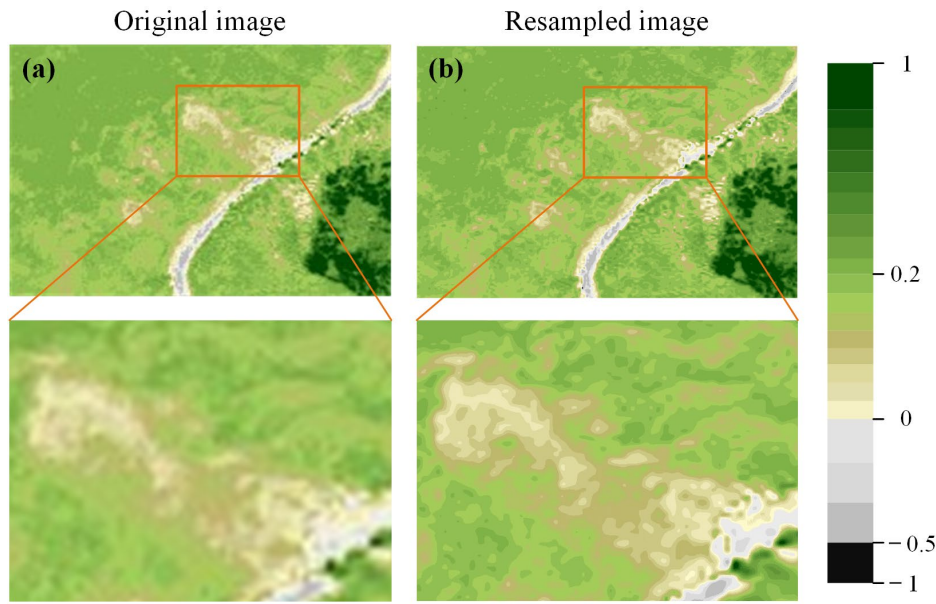

**Fig. 2.** Comparison of NDVI images: (a) Original 10 m resolution; (b) After resampling to 1 m resolution.

2.2.2 Stage 2: Constructing Video Data from Time-Series Imagery

To track the spatiotemporal evolution of landslide scars using a vision foundation model, preprocessed multi-temporal NDVI image sequences are converted into video sequences. Specifically, $n$ NDVI images are chronologically ordered and assigned sequential frame indices from 0 to $n-1$. All frames are uniformly converted to a standard image format (e.g., JPG or PNG) and compiled either into a single video file or as individual frames within a directory. This transformation converts discrete, static remote sensing datasets into continuous dynamic video data, enabling continuous tracking of landslide evolution.

2.2.3 Stage 3: Tracking Spatiotemporal Evolution of Landslide Scars from Video Data with a Vision Foundation Model

This study employs the Segment Anything Model 2 (SAM 2) for segmentation tasks (Fig. 3). SAM 2 is a vision foundation model designed for both image and video segmentation. It supports prompt-based segmentation, allowing precise delineation of targets using user-provided inputs such as points, bounding boxes, or masks (Ravi et al., 2024). A key advantage of SAM 2 is its built-in temporal memory module, which retains object information from previous frames. This capability enables the model to maintain segmentation continuity and iteratively refine results throughout the video sequence based on new prompt information.

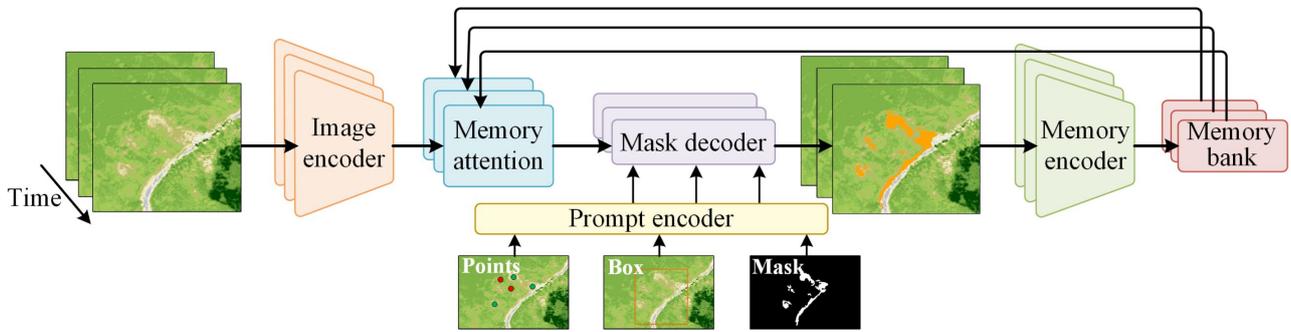

**Fig. 3.** Architecture of the employed vision foundation model (SAM 2).

The procedure of using SAM 2 to track landslide scars consists of the following four steps:

**(1) Initialization guided by domain knowledge:** In the first video frame (Frame 0), researchers provide a limited number of prompt points based on domain knowledge. Positive prompt points are placed within the landslide scar, while negative prompt points are positioned in the surrounding stable areas. SAM 2 then generates an initial segmentation mask of landslide scars based on these prompts.

**(2) Temporal propagation and automated tracking:** After generating the initial segmentation mask in the first video frame, SAM 2 utilizes its temporal memory and propagation mechanisms to automatically extend the mask to all subsequent frames. This process enables continuous tracking of the landslide scar boundary. The model dynamically adapts to changes in boundary extent and morphology, performing automated segmentation consistently across the entire time-series.

**(3) Interactive refinement and optimization:** During automated tracking, deviations may occur in one or more frames, including boundary blurring, region omissions, or over-segmentation. In such cases, researchers can provide additional positive or negative prompt points on the affected frames. SAM 2 immediately incorporates these corrections and utilizes its temporal memory to propagate the adjustments, thereby improving segmentation accuracy in subsequent frames and ensuring consistent reliability across the entire sequence (Fig. 1).

**(4) Spatiotemporal evolution feature extraction:** The model outputs a sequence of binary masks corresponding exactly to each input video frame. This sequence accurately represents the spatial extent and morphology of the landslide scar at every time step. Spatiotemporal analysis of the mask sequence allows quantitative characterization of dynamic features, including variations in area, expansion direction, and deformation rate. This process generates a high-precision dynamic sequence of landslide scars, providing a foundation for subsequent quantitative analyses.

**2.3 Verification of the Proposed Framework**

The performance of the proposed framework is validated through quantitative comparison with manually interpreted ground truth data. To evaluate applicability across different stages of landslide evolution, two representative scenarios were selected.

**(1) Post-failure landslides:** Historical image sequences covering the complete failure process were used to assess the model's ability to retrospectively reproduce the full dynamic evolution of landslide scars, from initial deformation through final morphology.

**(2) Active landslides:** Cases exhibiting continuous creep or progressive deformation were selected to evaluate the model's capability to capture dynamic changes in local landslide scars.

To quantitatively assess the performance of the proposed framework, three standard image segmentation metrics were employed: Intersection over Union (IoU), precision, and recall. These metrics were computed on a per-frame basis to evaluate the framework's ability to identify landslide scars dynamically throughout the entire time-series. Specifically, IoU measures the spatial overlap between the predicted landslide scars and the ground truth and is widely recognized as the primary metric in image segmentation tasks (Eq.2). Precision represents the fraction of pixels classified as landslide scars by the model that are correctly identified (Eq.3). Recall represents the fraction of true landslide scars pixels that are successfully detected by the model (Eq.4).

$$IoU = \frac{|A \cap B|}{|A \cup B|} \quad (2)$$

$$Precision = \frac{TP}{TP + FP} \quad (3)$$

$$Recall = \frac{TP}{TP + FN} \quad (4)$$

Here, $A$ denotes landslide scars predicted by the model, while $B$ represents the manually annotated ground truth. True Positive ($TP$) indicates pixels correctly identified as part of the landslide scar. False Positive ($FP$) refers to background pixels incorrectly classified as part of landslide scars, and False Negative (*FN*) denotes landslide scar pixels that were not detected. These metrics provide a standardized and rigorous framework for quantitatively evaluating the performance of the proposed method in dynamically identifying landslide scars.

## 3. Results and Analysis

### 3.1 Application Scenario 1: A Post-failure Landslide – Baige Landslide

3.1.1 Study Area and Data

The Baige landslide is located on the right bank of the Jinsha River, at the boundary between Jiangda County in the Tibet Autonomous Region and Baiyu County in Sichuan Province, China. The landslide experienced two large-scale failures on October 10 and November 3, 2018, with estimated volumes of approximately $2.3 \times 10^7$ m$^3$ and $8.5 \times 10^6$ m$^3$, respectively (Fan et al., 2019).

The study area is located in the Jinsha River basin within the Hengduan Mountains. The aspect ranges from 80° to 105°, with elevations between 2,880 and 3,720 m and a relative relief of approximately 840 m. The landslide measures approximately 1,330 m in length and 670 m in maximum width, with an average slope angle of 30° (Xiong et al., 2020). The dominant lithologies are gneiss and serpentinite, intersected by several northwest-trending faults, including the Boluo–Muxie Fault. The region receives an average annual precipitation of about 620 mm, predominantly between June and September. Vegetation exhibits pronounced seasonal variability, with moderate density in summer and sparse coverage during winter and spring (Fan et al., 2019; Xiong et al., 2020). Although a substantial portion of the loose material was removed during the two failure events, the slope remains structurally disturbed, with persistent instability along the rear edge and lateral flanks (Ma et al., 2025).

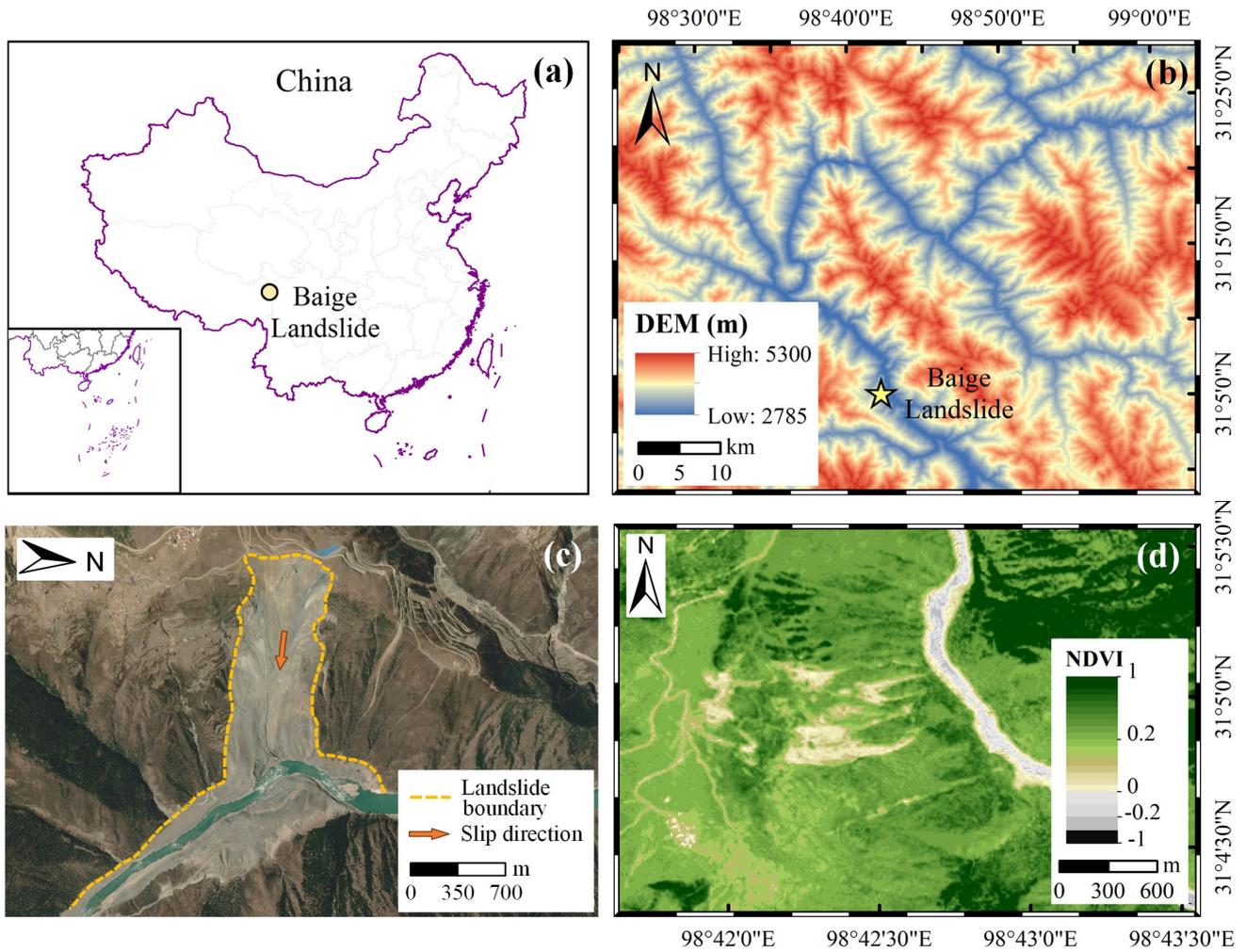

**Fig. 4.** The Baige landslide: (a) Geographic location; (b) Elevation distribution; (c) Landslide overview; (d) Sentinel-2 NDVI imagery from January 16, 2017.

To investigate the continuous evolution of the Baige landslide scars before and after failure, 156 cloud-free L2A products from the Sentinel-2 satellite acquired between January 2017 and June 2025 were used to generate NDVI images. The original imagery was resampled to an effective spatial resolution of 2 m. Landslide scars were delineated using an NDVI threshold of ≤ 0.1, with slightly elevated thresholds applied during periods of dense vegetation, such as summer. Monitoring these low-NDVI areas enabled quantitative characterization of the long-term deformation dynamics of the landslide.

3.1.2 Verification of the Proposed Framework for Tracking the Spatiotemporal Evolution of the Baige Landslide Scars

Comparison of the framework results with manual interpretation reveals high consistency in both spatiotemporal distribution and boundary delineation (Fig. 5). Frame-by-frame evaluation across the entire sequence indicates average IoU, precision, and recall values of 0.919, 0.963, and 0.952, respectively, demonstrating the framework's capability to capture the spatiotemporal evolution of landslide scars (Fig. 6). As an example, on January 16, 2017, the framework identified a landslide scar area of $2.32 \times 10^5$ m$^2$, whereas manual interpretation yielded $2.28 \times 10^5$ m$^2$, corresponding to a relative error of 1.75%. The IoU, precision, and recall for this frame were 0.807, 0.887, and 0.900, respectively. Prior to the large-scale failure on October 10, 2018, landslide scars were distributed discontinuously and fragmented, with IoU values fluctuating to a minimum of 0.72 (Fig. 6a). Over time, landslide scars gradually expanded and interconnected, resulting in improved identification accuracy and stable IoU values exceeding 0.9.

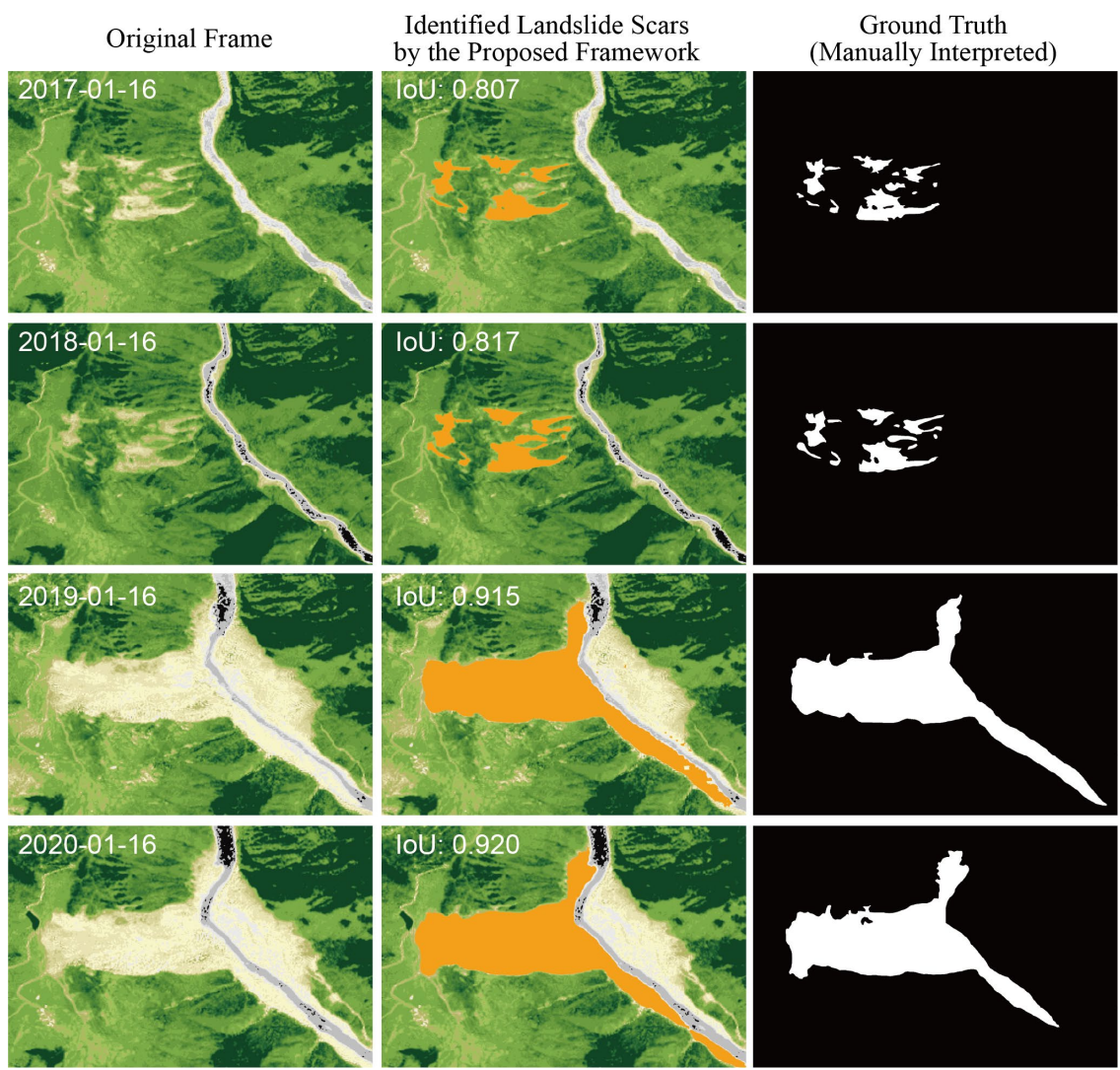

**Fig. 5.** Comparison between the identified landslide scars and the manually interpreted landslide scars (ground truth) of the Baige landslide.

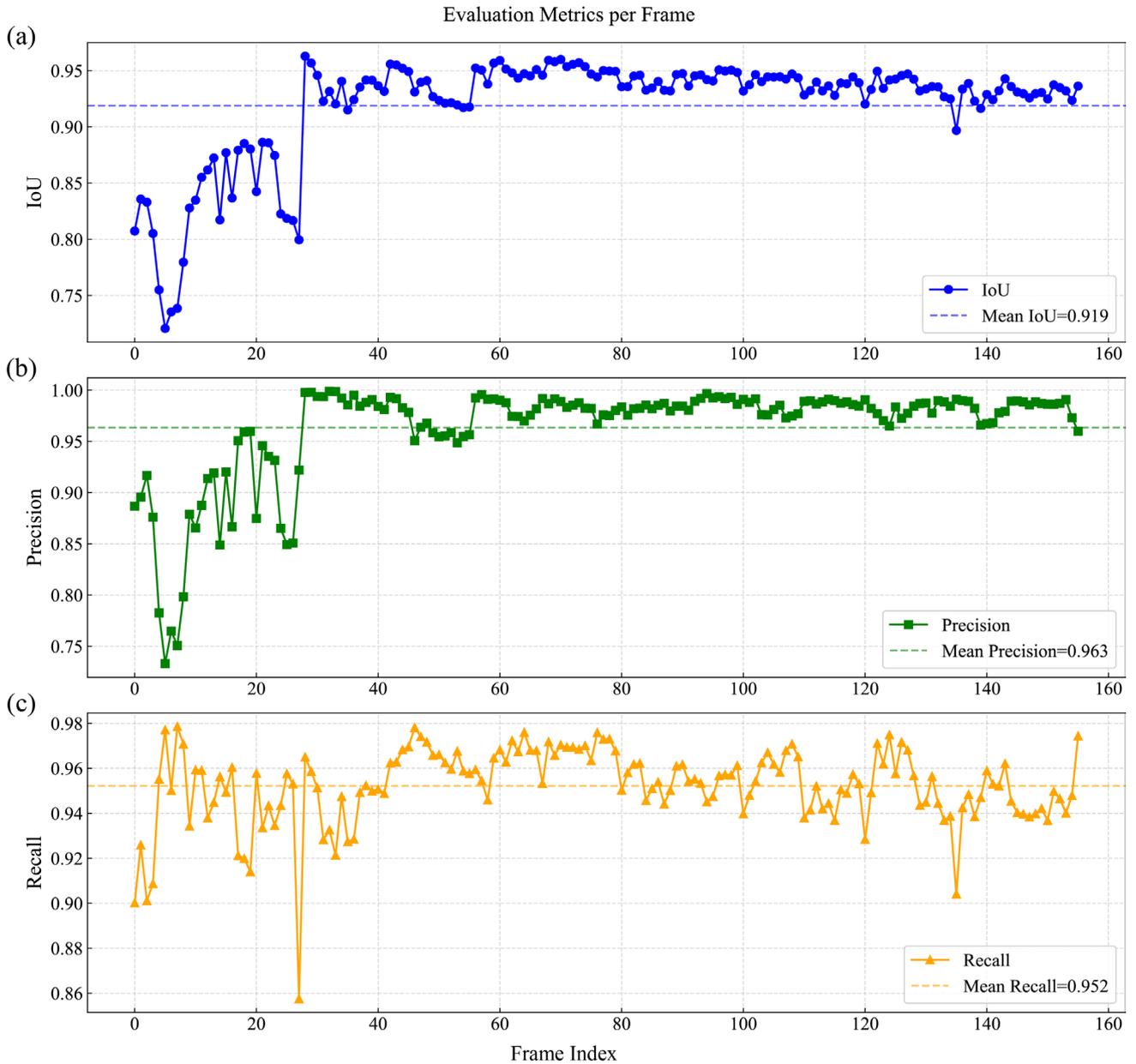

**Fig. 6.** Accuracy of the proposed framework applied to track the spatiotemporal evolution of the Baige landslide scars: (a) IoU; (b) Precision; (c) Recall.

3.1.3 Analysis of the Spatiotemporal Evolution of the Baige Landslide Scars

On the basis of the high-precision identification results, the complete dynamic evolution process of the Baige landslide before and after failure was further analyzed.

(1) Pre-failure Progressive Deformation Stage

The results indicate that, prior to failure, the Baige landslide underwent a dynamic process characterized by the gradual expansion and interconnection of localized, discontinuous fractured landslide scars (Fig. 7). To quantify this process, the temporal evolution of the landslide scar area was tracked (Fig. 8a). Although seasonal vegetation changes caused non-monotonic fluctuations, the overall trend exhibits a steady increase in the landslide scar area, from $2.32 \times 10^5$ m$^2$ on January 16, 2017, to $3.55 \times 10^5$ m$^2$ on August 9, 2018, corresponding to a 53% increase. Seasonal comparisons were conducted to minimize vegetation-related interference, confirming the progressive expansion of the creep rupture zone (Figs. 8b, 8c). Analysis indicates that the landslide entered an accelerated deformation phase beginning on November 7, 2017, with rapid acceleration observed after June 5, 2018. This sustained expansion reflects the large-scale precursor deformation process as the landslide transitioned from creep to accelerated deformation. The abrupt decrease in area on September 18, 2018, was attributed to the obscuring effect of cloud shadows.

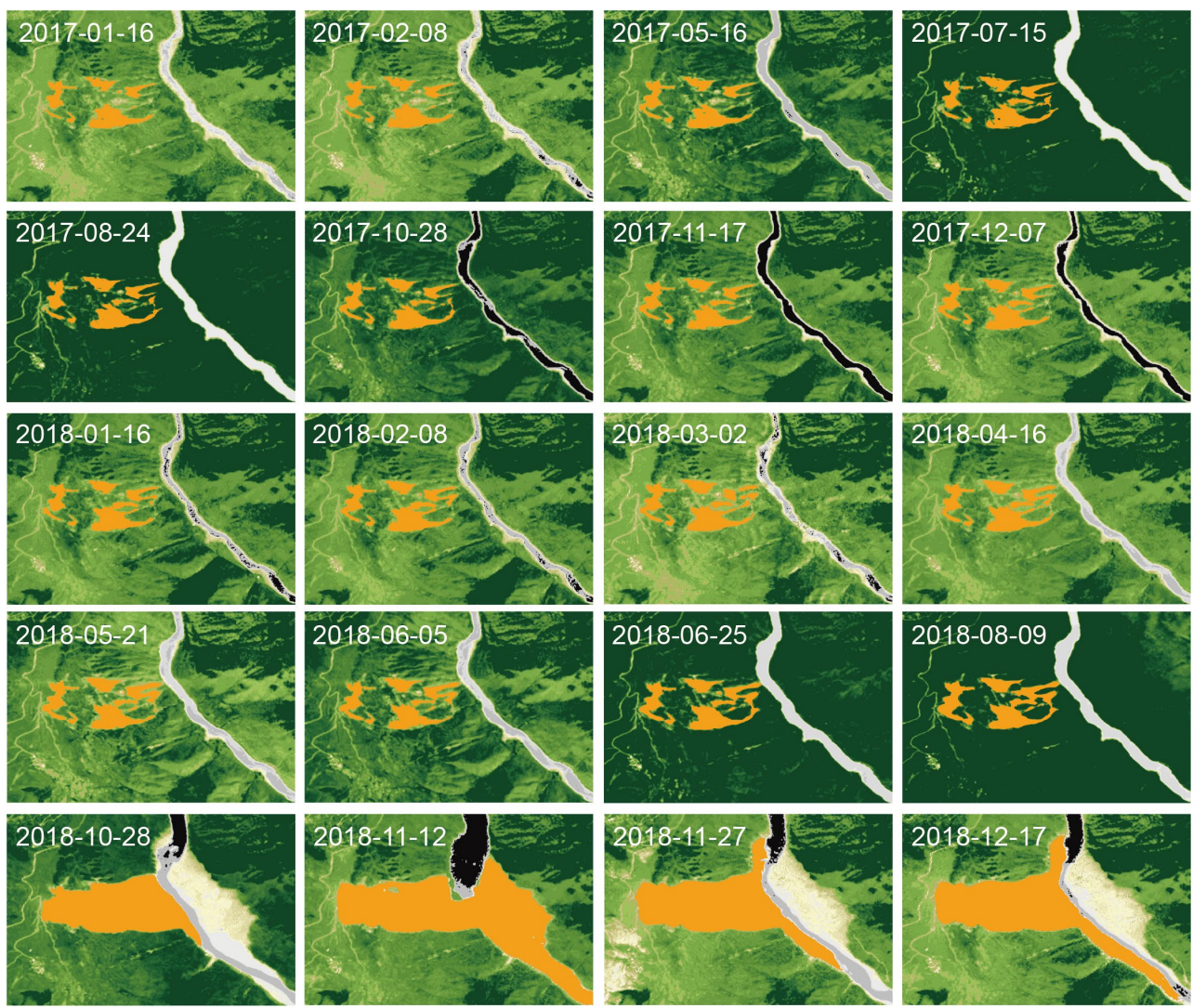

**Fig. 7.** Tracked spatiotemporal evolution of the Baige landslide scars from January 2017 to December 2018 (orange masks represent the identified landslide scars).

(2) Precise Capture of the Failure Events

The time-series analysis of the Baige landslide scar area identified two major failure events, evident as distinct spikes in the curve (Fig. 8a). These spikes correspond to the first failure on October 10, 2018, and the second on November 3, 2018. Following the second event, the identified area on November 12, 2018, showed a marked increase, resulting from the inclusion of both the landslide source area and the accumulation zone formed by the damming of the Jinsha River. By November 22, after the discharge of the landslide-dammed lake, the accumulation zone was removed, leading to a corresponding decrease in the mapped area.

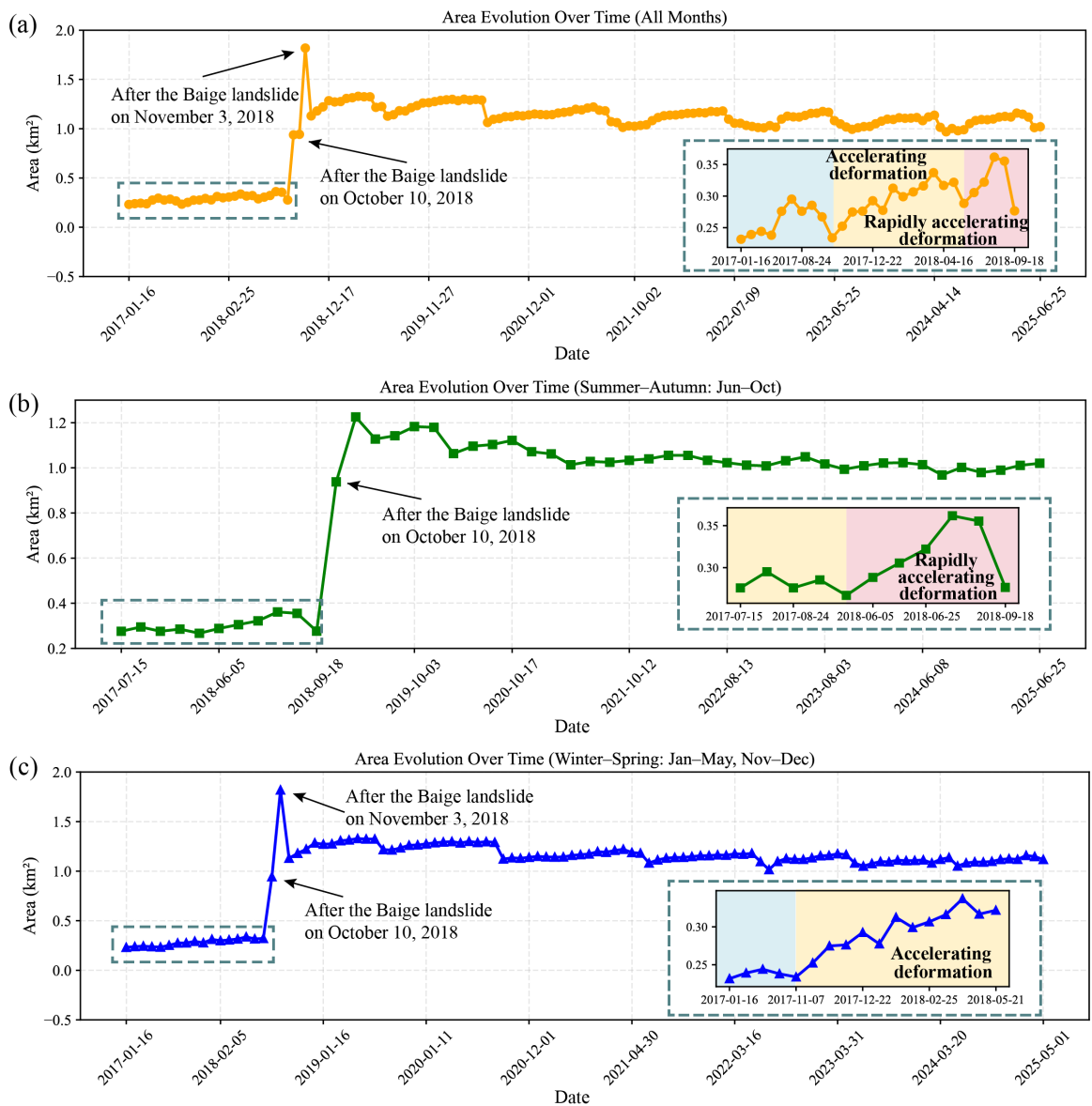

**Fig. 8.** Area changes of the Baige landslide scars: (a) Complete time-series from 2017 to 2025; (b) Area variations during summer and autumn (June–October); (c) Area variations during winter and spring (January–May and November–December)

(3) Post-failure Continuous Evolution Monitoring

Furthermore, the proposed framework effectively monitors the post-failure evolution of the landslide mass. While the overall landslide scar area has declined due to regional stabilization and vegetation recovery, residual activity remains in the unstable rear sector. Interannual comparison of landslide scar boundaries reveals notable expansion in this sector (Fig. 9). For instance, the boundary on November 27, 2019, expanded substantially relative to the same date in 2018 (Fig. 9a), and a new expansion zone appeared north of the trailing edge after November 11, 2023 (Fig. 9e). These new expansion zones correspond to the K1 and K2 unstable zones at the trailing edge of the Baige landslide, consistent with previous findings (Fan et al., 2019; Xiong et al., 2020; Ma and Mei, 2025), thereby validating the framework's utility for long-term post-failure stability monitoring.

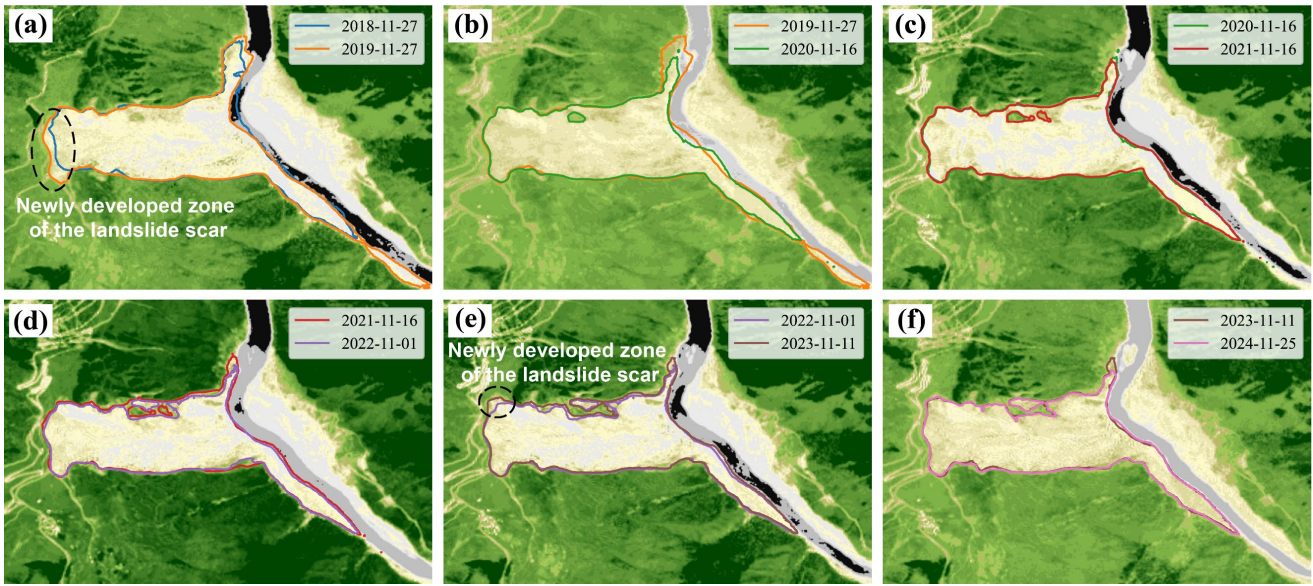

**Fig. 9.** Boundaries of the Baige landslide scar in corresponding periods across different years.

**3.2 Application Scenario 2: An Active Landslide – Sela Landslide**

3.2.1 Study Area and Data

The Sela landslide is situated on the right bank of the Jinsha River in Mindu Township, Gongjue County, Tibet Autonomous Region, China (Fig. 10a). The landslide exhibits a tongue-shaped planform, with a longitudinal extent of 1,280 ~ 1,551 m, a transverse width of 986 ~ 1,046 m, a total area of approximately $1.63 \times 10^6$ m$^2$, and an estimated volume of $6.52 \times 10^7$ m$^3$ (Fig. 10c). Crest and toe elevations are 3,342 m and 2,649 m, respectively, corresponding to a maximum vertical relief of about 693 m. Topographically, the slope is characterized by steep upper sectors and gentler lower sectors, with mean slope angles of 30° ~ 35° and local slopes reaching 40° ~ 45° (Zhu et al., 2021).

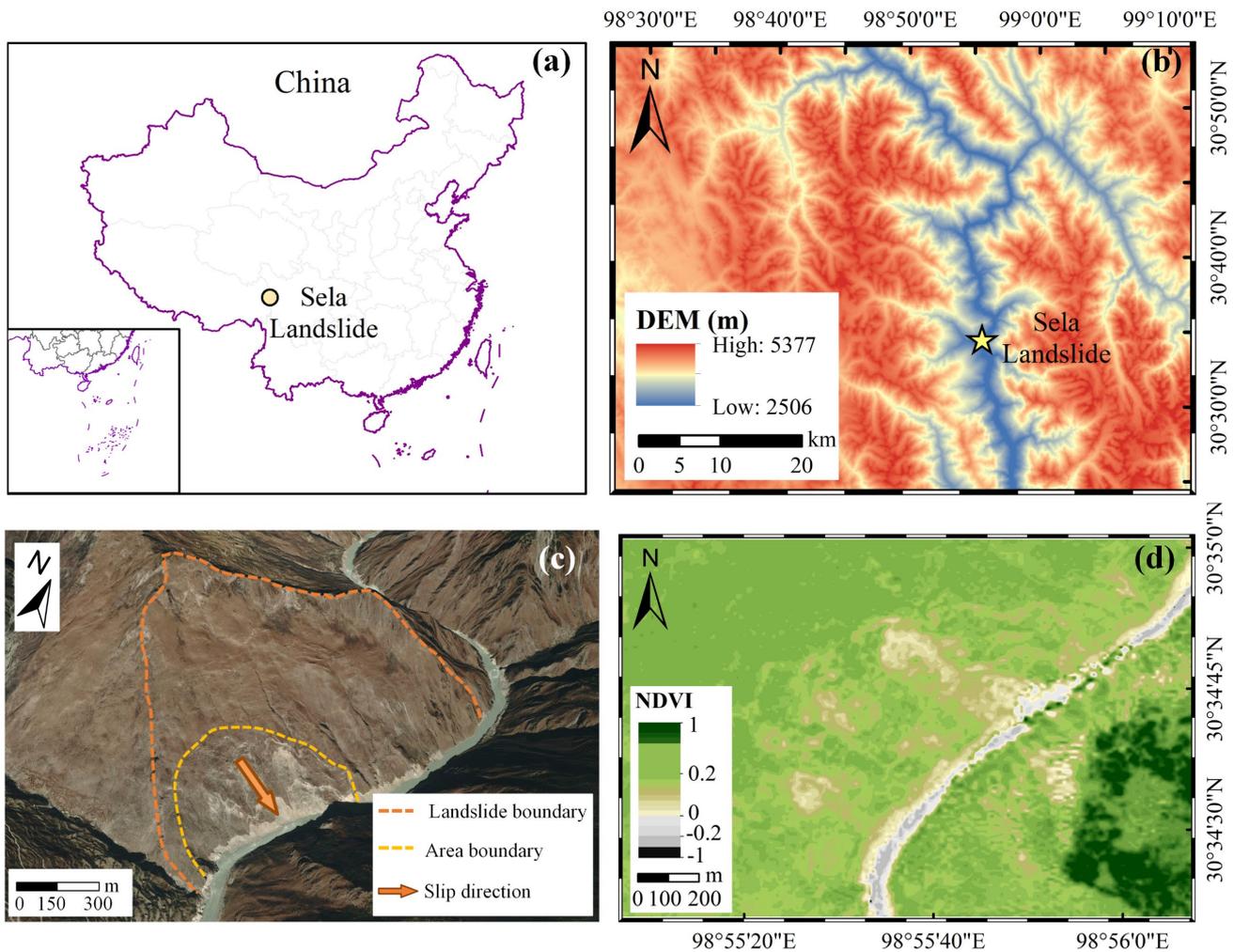

**Fig. 10.** The Sela landslide: (a) Geographic location; (b) Elevation distribution; (c) Landslide overview; (d) Sentinel-2 NDVI imagery from February 12, 2017.

Structurally, the landslide is situated within the Jinsha River suture zone. Regional faulting is highly developed due to the eastward thrusting of the Tibetan Plateau, resulting in a fractured and weak rock mass. The middle and rear sections exhibit numerous tensile fractures and gully erosion, while the frontal slope shows pronounced disintegration. In addition, continuous undercutting by the Jinsha River at the toe has formed cliffs 30 ~ 150 m high (Zhu et al., 2021).

Yan et al. (2024), integrating optical remote sensing and multi-temporal SAR observations, report that the Sela landslide is undergoing sustained creep deformation. The creep rate increased markedly after the two Baige failure events in 2018. Flooding associated with those failures caused intense toe scouring at the Sela landslide, which in turn promoted toe instability and scour-induced retrogressive deformation. Consequently, the slope remains susceptible to catastrophic failure under extreme rainfall or strong seismic loading. Such a failure could trigger a cascade of hazards, including river damming, breach-induced outburst flooding, and downstream inundation, thereby posing significant risks to downstream hydropower facilities, bridges, settlements, and other riverside infrastructure.

To monitor the spatiotemporal evolution of the Sela landslide scar, 91 cloud-free L2A scenes from the Sentinel-2 satellite acquired between February 2017 and June 2025 were used to compute NDVI time-series. The original images were resampled to an effective spatial resolution of 1 m (Fig. 10d). Areas with NDVI $\leq 0.1$ (with slightly elevated thresholds applied during periods of vigorous vegetation growth) were classified as bare ground and interpreted as exposed surfaces resulting from deformation, fissuring, or collapse. Temporal tracking of these low-NDVI areas was used to characterize the landslide's dynamic evolution.

3.2.2 Verification of the Proposed Framework for Tracking the Spatiotemporal Evolution of the Sela Landslide Scars

To evaluate the framework's reliability on active landslides, the automated identification of the Sela landslide was compared with results from manual interpretation (Fig. 11). The two approaches showed strong agreement in both the spatiotemporal distribution and boundary delineation of landslide scars. Quantitative assessment revealed average IoU, precision, and recall values of 0.774, 0.877, and 0.873, respectively, during 2017–2025 (Fig. 12). Although these values are slightly lower than those obtained for the Baige landslide, they still confirm the framework's robustness in tracking the spatiotemporal evolution of persistent creep landslide scars. The minor reduction in accuracy is likely related to the fragmented nature of local scars in active landslides and the difficulty of delineating their indistinct boundaries.

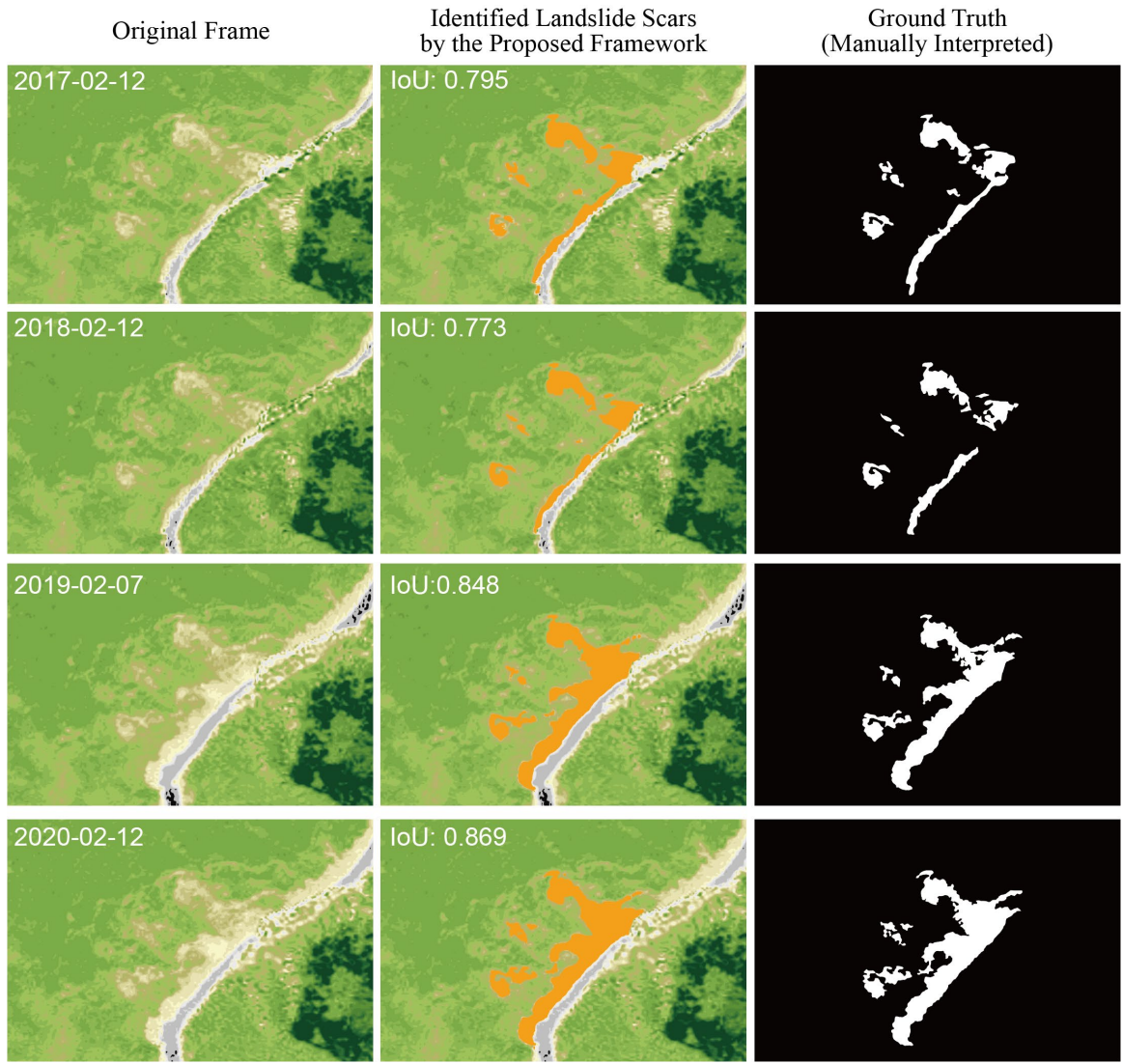

**Fig. 11.** Comparison between the identified landslide scars and the manually interpreted landslide scars (ground truth) of the Sela landslide.

3.2.3 Analysis of the Spatiotemporal Evolution of the Sela Landslide Scars

A continuous deformation record of the Sela landslide spanning February 2017 to June 2025 was reconstructed using the proposed framework; the early-stage spatiotemporal evolution (February 2017 to December 2018) is shown in Fig. 13.

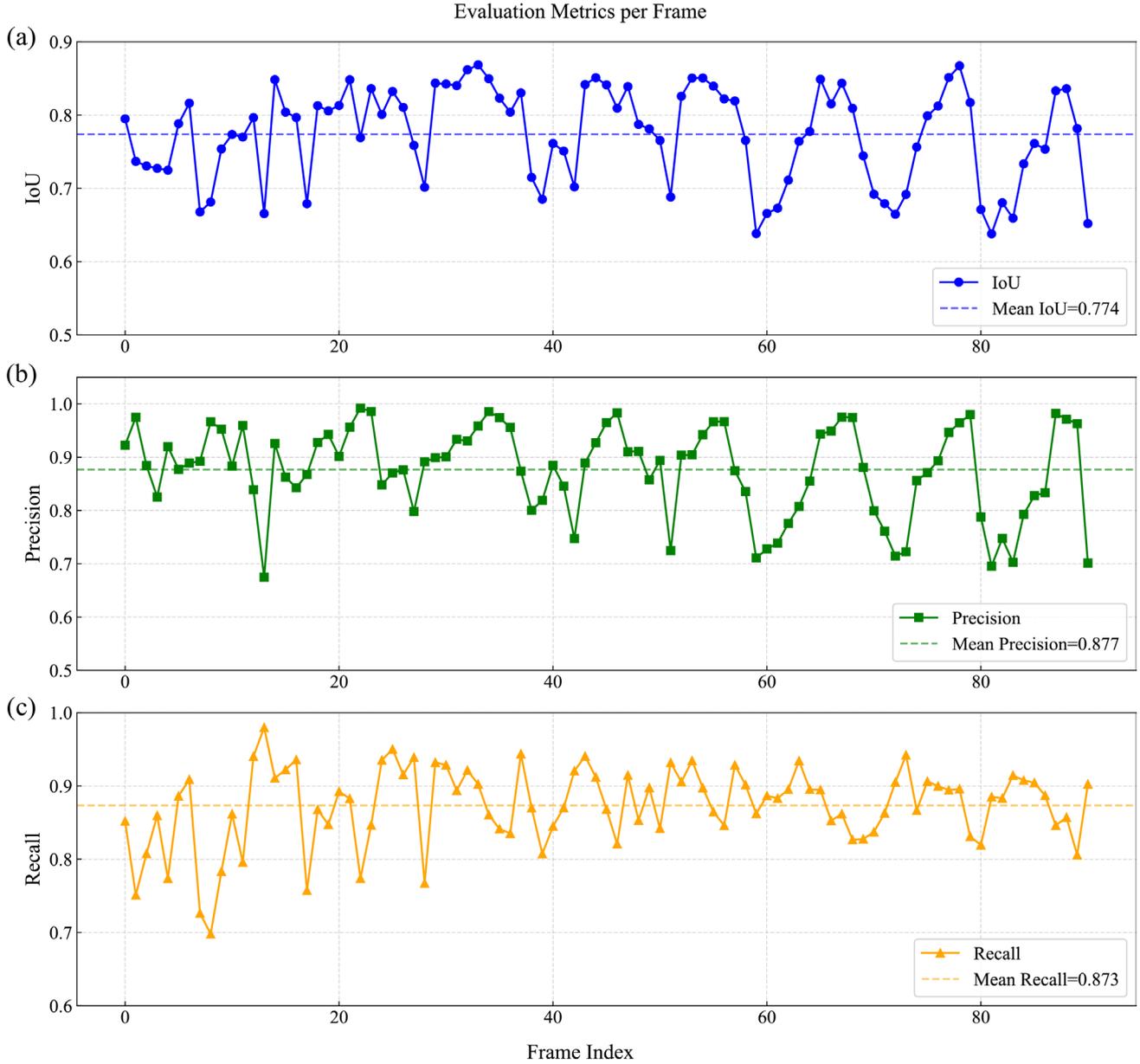

**Fig. 12.** Accuracy of the proposed framework applied to track the spatiotemporal evolution of the Sela landslide scars: (a) IoU; (b) Precision; (c) Recall.

(1) Response to External Triggering Events

The identification results reveal a clear turning point in the landslide evolution occurring in November 2018 (Fig. 14). Before this date, the landslide scar remained limited in area and exhibited slow growth; afterwards, the area expanded rapidly and subsequently displayed pronounced fluctuations. The most rapid expansion occurred following the Baige landslide breach flood on November 3, 2018, with the deformation-zone area increasing from $2.823 \times 10^4$ m$^2$ (October 5, 2018) to $1.212 \times 10^5$ m$^2$ (November 19, 2018), corresponding to an approximately 330% increase.

Comparison of the boundaries identified on November 19, 2018 with those of October 5, 2018 indicates that the expansion was concentrated within the riverbank erosion zone (Fig. 13). This abrupt change is temporally coincident with the extreme external disturbance, and the evidence is consistent with the hypothesis that intense scouring at the slope toe accelerated landslide deformation.

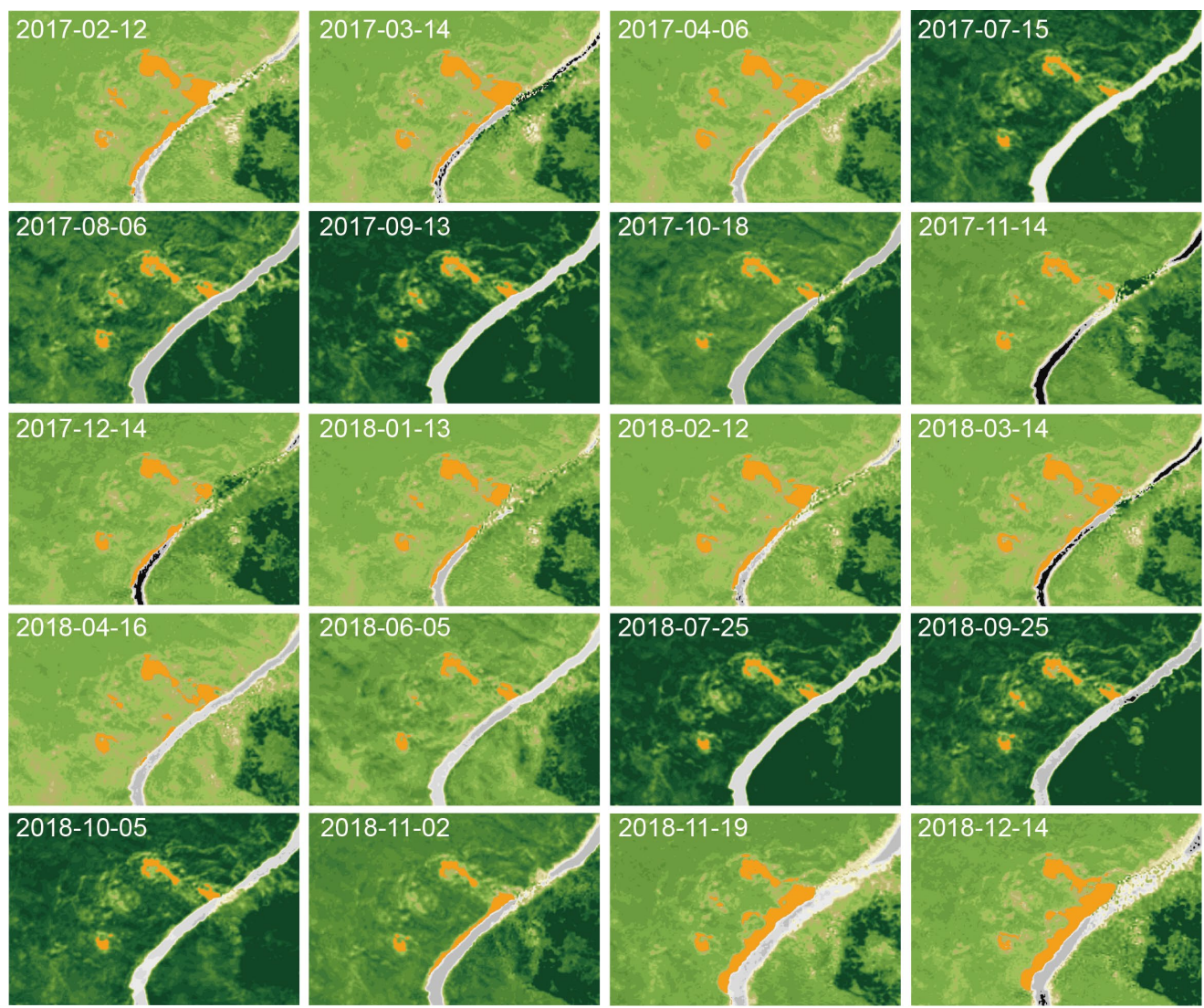

**Fig. 13.** Tracked spatiotemporal evolution of the Sela landslide scars from February 2017 to December 2018 (orange masks represent the identified landslide scars).

(2) Analysis of Periodic Fluctuations and Seasonal Effects

Over the long-term evolution, the area of the landslide scars shows pronounced seasonal fluctuations, with minima typically occurring in summer when vegetation growth is most vigorous (Fig. 14a). These fluctuations reflect the influence of periodic vegetation dynamics on NDVI-based identification. Even after accounting for seasonal effects, residual variations remain evident in the

summer–autumn and winter–spring subsections (Figs. 14b, 14c). Interannual comparison reveals that, relative to November 2017, the landslide scar boundaries expanded markedly in November 2018 and November 2019, but began to contract after 2020 (Fig. 15). These findings indicate that the Sela landslide does not exhibit a steadily accelerating trend, but rather alternates between phases of activation and stabilization.

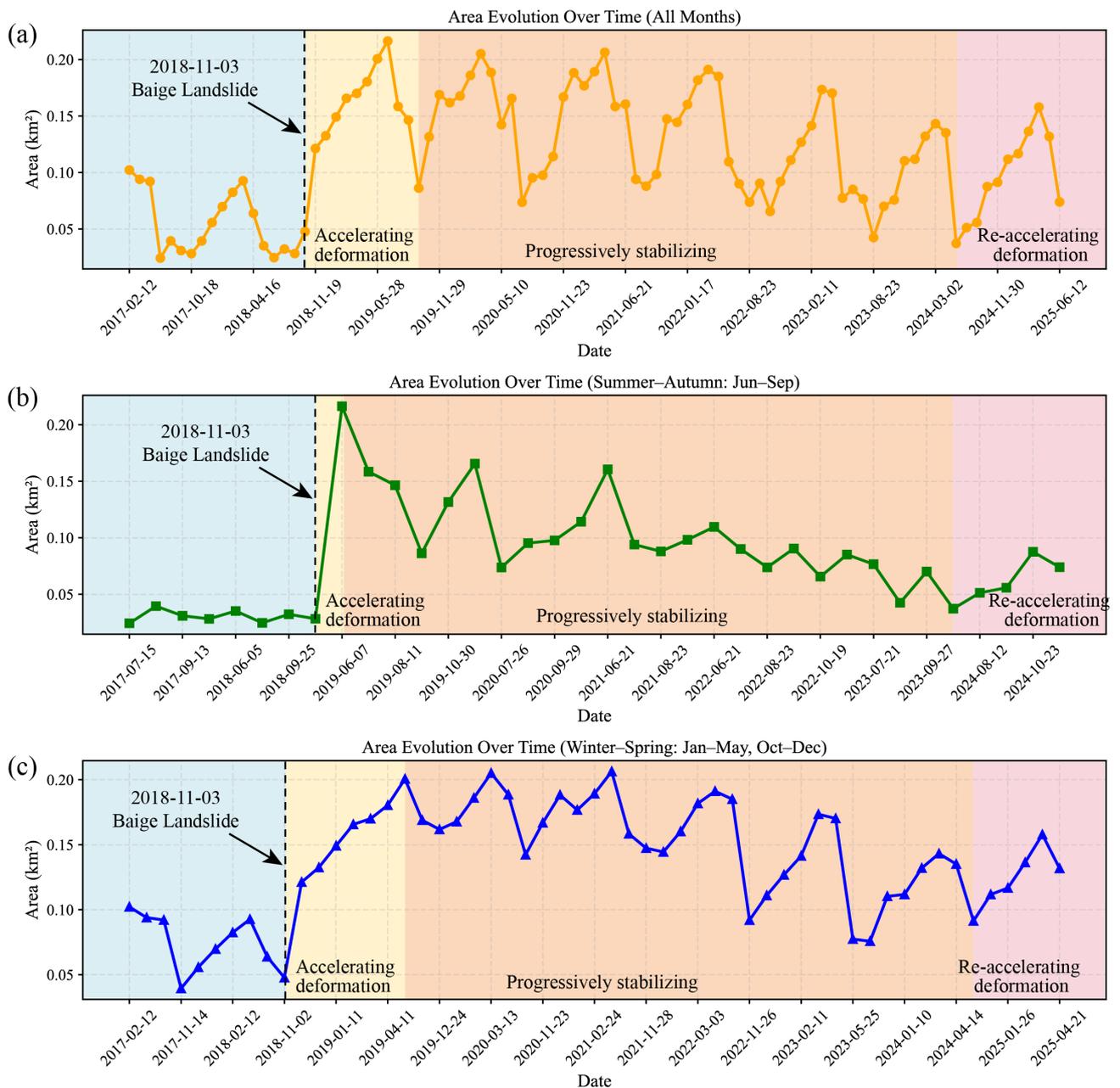

**Fig. 14.** Area changes of the Sela landslide scars: (a) Complete time-series from 2017 to 2025; (b) Area variations during summer and autumn (June–October); (c) Area variations during winter and spring (January–May and November–December).

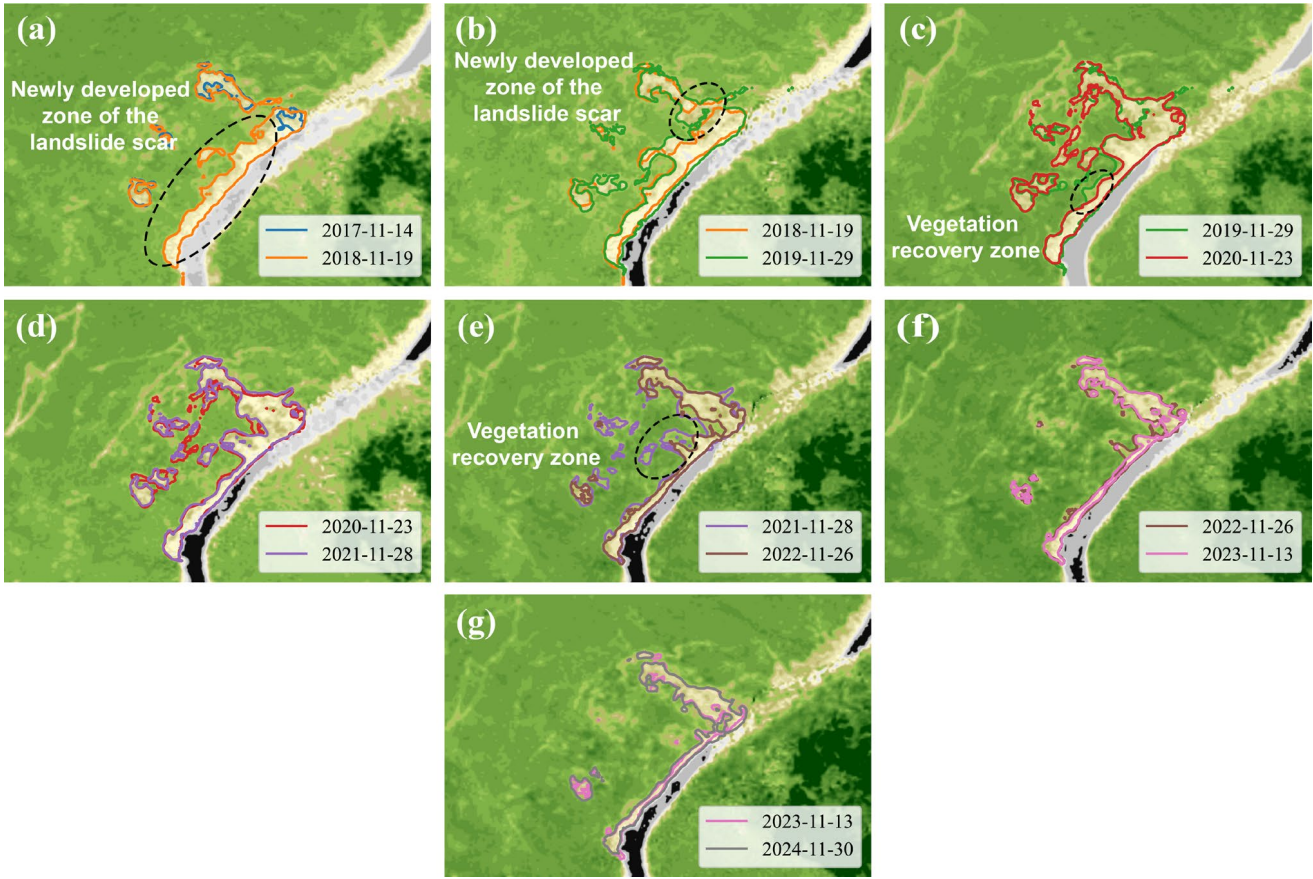

**Fig. 15.** Boundaries of the Sela landslide scar in corresponding periods across different years.

(3) Long-term Trend and Landslide Activity Assessment

Despite seasonal fluctuations, inter-seasonal comparison of long-term trends reveals the underlying patterns of landslide deformation. Analysis of the summer–autumn and winter–spring sub-sequences allows the following observations (Fig. 14):

Post-flood stress adjustment: Following the 2018 flood event, the landslide scar expanded markedly and continued to grow over the subsequent six months, through June 2019. This expansion may be associated with stress redistribution at the disturbed slope front.

Mid-term relative stability: From the second half of 2019, the landslide scar area gradually decreased and stabilized, partly reflecting ecological recovery of vegetation within the disturbed zones.

Recent reactivation: Notably, the landslide scar exhibited renewed expansion during 2024–2025. This observation indicates that the Sela landslide remains active and underscores the necessity for continued monitoring.

## 4. Discussion

### 4.1 Motivation and Novelty of the Proposed Framework

The evolution of landslide scars is inherently continuous. Accurate tracking of the spatiotemporal evolution of landslide scars is critical for deeply understanding failure mechanisms, detecting precursors, and assessing post-failure stability. However, most existing landslide detection approaches, including deep learning models, treat multi-temporal remote sensing image analysis as a series of independent tasks (Liu et al., 2021; Lv et al., 2023; Wan et al., 2023). This limitation constrains their capacity for continuous identification of landslide scars. To address this challenge, this study reconstructs discrete multi-temporal remote sensing image sequences into continuous video data, enabling the use of vision foundation models originally developed for video segmentation. The proposed framework facilitates continuous monitoring of landslide deformation, capturing both progressive pre-failure deformation crucial for early-warning applications and post-failure evolution necessary for evaluating secondary hazards and long-term stability.

The proposed framework captures and utilizes temporal information in long-term image sequences. Instead of performing segmentation independently on each frame, the framework leverages its temporal memory mechanism to infer kinematic relationships of landslide scar boundaries between adjacent frames. This is crucial for understanding landslide evolution from localized creep and progressive deformation to catastrophic failure, followed by post-failure evolution. Case studies show that the framework can accurately track the evolution of landslide scars, even with 10 m resolution optical imagery. This demonstrates its effectiveness and robustness in capturing macroscopic deformation dynamics.

Furthermore, compared to existing supervised deep learning models, such as CNNs, the proposed framework reduces dependence on large-scale pixel-level annotated training data, enabling zero-shot recognition of landslide scars. Traditional methods require extensive data annotation and model training for specific scenarios, resulting in high costs and limited model generalization (Wan et al., 2023; Wu et al., 2024). In contrast, the proposed framework requires only a small number of interactive prompt points provided on the first frame to automatically complete tracking of the entire sequence, substantially reducing the application threshold and requirements for data preprocessing.

## 4.2 Time-Series Data Requirements for the Proposed Framework

The key novelty of the proposed framework lies in reconstructing multi-temporal remote sensing image sequences into continuous video sequences for subsequent analysis. Therefore, the primary prerequisite is acquiring optical remote sensing images with continuous time-series and precise registration. Because optical images are susceptible to interference from cloud cover, vegetation, and lighting conditions, ensuring the usability of each temporal image presents significant challenges. Consequently, the available time-series images are often non-equidistant (Lan et al., 2022; Casagli et al., 2023). The framework enables the construction of video data from images acquired at irregular time intervals. However, it should be noted that excessively long intervals between images may lead to discontinuities in the identification results.

Moreover, the spatial resolution of remote sensing data governs the level of detail in the identification results. High-resolution imagery can capture finer precursory deformation features, whereas medium-to-low resolution imagery (e.g., the 10 m resolution data used in this study) is more suitable for monitoring macro-scale deformation processes (Li et al., 2023). For medium-to-low resolution imagery, resampling techniques are applied to remove jagged edges, thereby improving the smoothness of the identification results.

It should also be noted that the susceptibility of optical imagery to interference from clouds, vegetation cover, and illumination conditions is an inherent characteristic of optical remote sensing, rather than a limitation specific to the proposed framework. Therefore, the application of this framework requires careful planning in data source selection and preprocessing workflows to ensure the production of high-quality landslide deformation videos. Future research will focus on integrating this framework with multi-source data (e.g., InSAR-derived continuous deformation fields, LiDAR point clouds) to construct more robust multimodal videos, thereby mitigating the inherent limitations of optical-only data (Ma and Mei, 2021; Yang et al., 2025).

## 4.3 Implications of Domain Knowledge in the Proposed Framework

A key feature of the proposed framework is the integration of domain knowledge to ensure both the accuracy and interpretability of the identification results. In the initial frame of the landslide videos, positive and negative prompt points should be defined based on expert judgment of geological and topographical features to establish a reliable segmentation baseline for the model. These prompt points should encompass typical landslide scars and stable zones, with their number and distribution determined by the image resolution and research objectives.

The interpretation of identification results requires geological plausibility checks. For instance, determining whether the model-generated masks correspond to exposed surfaces or landslide deposits should be guided by domain knowledge, while spatial consistency can be evaluated using DEM-derived slopes, topographic positions, or InSAR deformation fields to reduce subjectivity.

Furthermore, the incorporation of domain knowledge enables the model to maintain identification consistency under complex terrain conditions, such as non-landslide areas with similar spectral or textural characteristics. When identification results include low-confidence frames or exhibit substantial discrepancies with external data, researchers can intervene by adjusting masks or introducing additional prompt points. These corrections are then propagated to subsequent frames through the model's temporal mechanism, thereby improving overall accuracy and stability while minimizing manual intervention. Thus, domain knowledge not only provides the foundation for the effective operation of the framework but also offers a reliable basis for its application in disaster monitoring and early warning.

**4.4 Limitations and Future Directions**

Although the proposed framework demonstrates strong applicability and robustness in identifying the spatiotemporal evolution of landslide scars, further refinement is still needed. Current vision foundation models are primarily pre-trained on general datasets and lack specialized representation of geoscience features such as landslide fractures and debris textures, which may result in suboptimal identification performance. Future work could employ parameter-efficient fine-tuning (PEFT) methods, such as Low-Rank Adaptation (LoRA), to achieve domain-specific adaptation with limited geoscience datasets, thereby enhancing the model's sensitivity and capacity to capture landslide evolution (Houlsby et al., 2019; Zaken et al., 2021; Yu et al., 2024; Guo et al., 2025; Hou et al., 2025). Moreover, although the proposed framework has yielded high-precision results in representative cases, its applicability in diverse geological environments, including densely vegetated areas or rapidly evolving landslides under extreme climatic conditions, requires further validation. Future research will therefore focus on more case studies and assess the framework's generalization potential through multi-source data fusion and cross-scenario testing.

## 5. Conclusion

This study proposes a novel and universal framework for tracking the spatiotemporal evolution of large-scale landslide scars using a vision foundation model. The key idea behind the proposed framework involves reconstructing discrete time-series optical images into continuous video sequences and leveraging vision foundation models with temporal memory mechanisms to capture the dynamic evolution of landslide scars. The framework consists of three main steps: (1) acquisition and resampling of temporal optical images; (2) reconstruction of discrete temporal optical images into continuous video sequences; and (3) identification of landslide scars using a vision foundation model integrated with domain knowledge. The proposed framework was validated and applied to both previously failed and active landslides. The following conclusions can be drawn:

(1) The proposed framework reconstructs discrete optical images into continuous video sequences, enabling accurate tracking of the spatiotemporal evolution of landslide scars.

(2) The proposed framework can identify progressive deformation characteristics preceding landslide initiation by tracking the gradual expansion of exposed ground surfaces. It also supports continuous monitoring of post-failure deformation, including the further expansion of exposed ground surfaces and indications of renewed instability.

(3) The proposed framework is simple, effective, and generalizable.

**Disclosure of interest.**

The authors report there are no competing interests to declare

**Acknowledgments**

This research was jointly supported by the National Natural Science Foundation of China (Grants 42507245 and 42277161). Thanks to the European Space Agency (ESA) for providing Sentinel 2 image data support. The authors would like to thank the editor and the reviewers for their contributions.